\documentclass[preprints,article,accept,pdftex,moreauthors]{Definitions/mdpi} 
\firstpage{1} 
\pubvolume{1}
\issuenum{1}
\articlenumber{0}
\pubyear{2025}
\copyrightyear{2025}
\datereceived{ } 
\dateaccepted{ } 
\datepublished{ }
\hreflink{https://doi.org/} 

\Title{Mobile Coverage Analysis using Crowdsourced Geolocation Data}

\TitleCitation{Mobile Coverage Analysis using Crowdsourced Geolocation Data}

\Author{Timothy Wong $^{1}$\orcidA{}, Tom Freeman $^{1}$\orcidB{} and Joseph Feehily $^{1}$\orcidC{}}

\AuthorNames{Timothy Wong, Tom Freeman and Joseph Feehily}

\AuthorCitation{Wong, T.; Freeman, T.; Feehily, J.}

\address{%
$^{1}$ \quad Vodafone Group PLC\\
}

\corres{Correspondence: tom.freeman@vodafone.com, joseph.feehily@vodafone.com}

\abstract{
	Effective assessment of mobile network coverage and the precise identification of service weak spots are paramount for network operators striving to enhance user Quality of Experience (QoE). This paper presents a novel framework for mobile coverage and weak spot analysis utilising crowdsourced QoE data. The core of our methodology analyses coverage at the individual cell (antenna) level using empirical geolocation data before aggregating to the site level. A key contribution of this research is the application of the One-Class Support Vector Machine (OC-SVM) algorithm to calculate mobile network coverage. This approach models the decision hyperplane as the effective coverage contour, facilitating robust determination of coverage areas for individual cells and entire sites. The same methodology is extended to analyse crowdsourced service loss reports, thereby identifying and quantifying geographically localised weak spots. Our findings demonstrate the efficacy of this framework in accurately mapping mobile coverage and, crucially, in highlighting granular areas of signal deficiency, particularly within complex urban environments.
}


\begin{document}

\section{Background}

The empirical assessment of mobile network coverage predominantly relies on two distinct methodologies: traditional drive testing and contemporary crowdsourced data analysis. Drive testing, a long-established approach, involves systematic data collection utilising specialised Radio Frequency (RF) measurement equipment. As exemplified by regulatory bodies like the UK's Office of Communications (Ofcom) \cite{ofcom_drive_test}, this often includes sophisticated scanning receivers and calibrated devices capable of concurrently measuring key performance indicators across multiple mobile network operators and technologies (e.g., 2G, 3G, 4G, and 5G) along predefined routes \cite{ofcom_drive_route}. This method affords a high degree of control over testing parameters and precise geolocation, enabling the acquisition of granular data essential for ensuring regulatory compliance and informing policy-making. The primary strengths of drive testing lie in its precision, the depth of diagnostic information obtainable, and the ability to perform controlled, repeatable measurements. However, its operational costs are substantial due to specialised equipment and personnel, and its spatio-temporal coverage is inherently limited by logistical constraints, often restricting data to major transport corridors.

In contrast, crowdsourced data analysis leverages measurements collected from mobile apps on a multitude of consumer smartphones. This approach offers unparalleled scalability, providing extensive geographic and temporal coverage at a significantly lower operational cost than drive testing. The data typically include network coverage, internet connectivity, device information, and contextual metadata. Crowdsourced apps in particular enable extensive sampling in indoor locations that are often inaccessible to conventional drive testing. Notably, they also record instances of no signal, immediately logging a location when a phone has no coverage. These "no-service" data points highlight complete coverage holes that traditional methods might miss. Analysing these areas may reveal important socioeconomic patterns \cite{Koutroumpis2016} to drive discussions on evidence-based policymaking processes.

However, analysing crowdsourced geolocation data to pinpoint coverage gaps is non-trivial. This is due to variability in data quality caused by device heterogeneity and calibration differences, GPS inaccuracies, a lack of precise control over the measurement environment (e.g., distinguishing indoor from outdoor measurements), user behaviour, and potential sampling biases toward more populated areas or specific user demographics. 

Using crowdsourced data, we first contrast a geometric approach for signal coverage analysis with a machine-learning method. We then outline a time-series-based validation strategy and examine the results. Finally, we consider practical engineering applications of improved coverage analysis and their business impact.

\section{Methodology}

One traditional approach to estimating mobile coverage areas from crowdsourced data is to use computational geometry, for example by constructing a convex hull around all the geolocated points where service was observed. The convex hull provides a simple approximate boundary of the network's coverage. If a geographic area lies outside this hull, it can be flagged as a potential coverage hole. This method is conceptually straightforward and computationally efficient. Yet, the convex hull method tends to overgeneralise because it cannot represent concavities, internal holes, or irregular coverage boundaries and it is strongly affected by outliers or widely spaced points \cite{Neidhardt2013, Brunello2013}.

In our approach, we model coverage estimation as a one-class classification (novelty-detection) problem: given geolocated measurements that evidence usable service (the "inliers"), we learn the support of that distribution and treat points outside it as likely weak or no coverage. The One-Class SVM (OC-SVM) algorithm \cite{10.1162/089976601750264965} estimates a function \(f(x)\) whose sign indicates membership in the learned support. The prediction \(f(x)\ge0\) denotes coverage, whereas \(f(x)<0\) indicates a lack of coverage. Unlike convex geometric baselines, the kernelised decision boundary can be highly non-convex, capturing concavities and internal holes driven by terrain, clutter, or shadowing. In practice, we use the Radial Basis Function (RBF) kernel to obtain smooth, locality-aware boundaries that wrap around dense regions of positive evidence without being pulled by isolated outliers. This mirrors common use of one-class SVM in anomaly detection, where the algorithm learns the region occupied by "normal" data and flags deviations as anomalies.

A key advantage of OC-SVM is the soft boundary controlled by the hyperparameter \(\nu \in (0,1]\). The \(\nu\) parameter simultaneously sets an upper bound on the fraction of training errors (points allowed outside the learned support) and a lower bound on the fraction of support vectors. It controls the trade-off between overfitting (making the boundary too tight, excluding real coverage) and underfitting (making the boundary too loose, including false coverage). A small \(\nu\) yields a tighter boundary that may exclude some true coverage points, while a large \(\nu\) produces a looser boundary that may include false positives. Thus, \(\nu\) governs the model's sensitivity to outliers and the complexity of the learned support region.

\begin{figure}[H]
\begin{adjustwidth}{-\extralength}{0cm}
\centering
\subfloat[\centering]{\includegraphics[width=5.0cm]{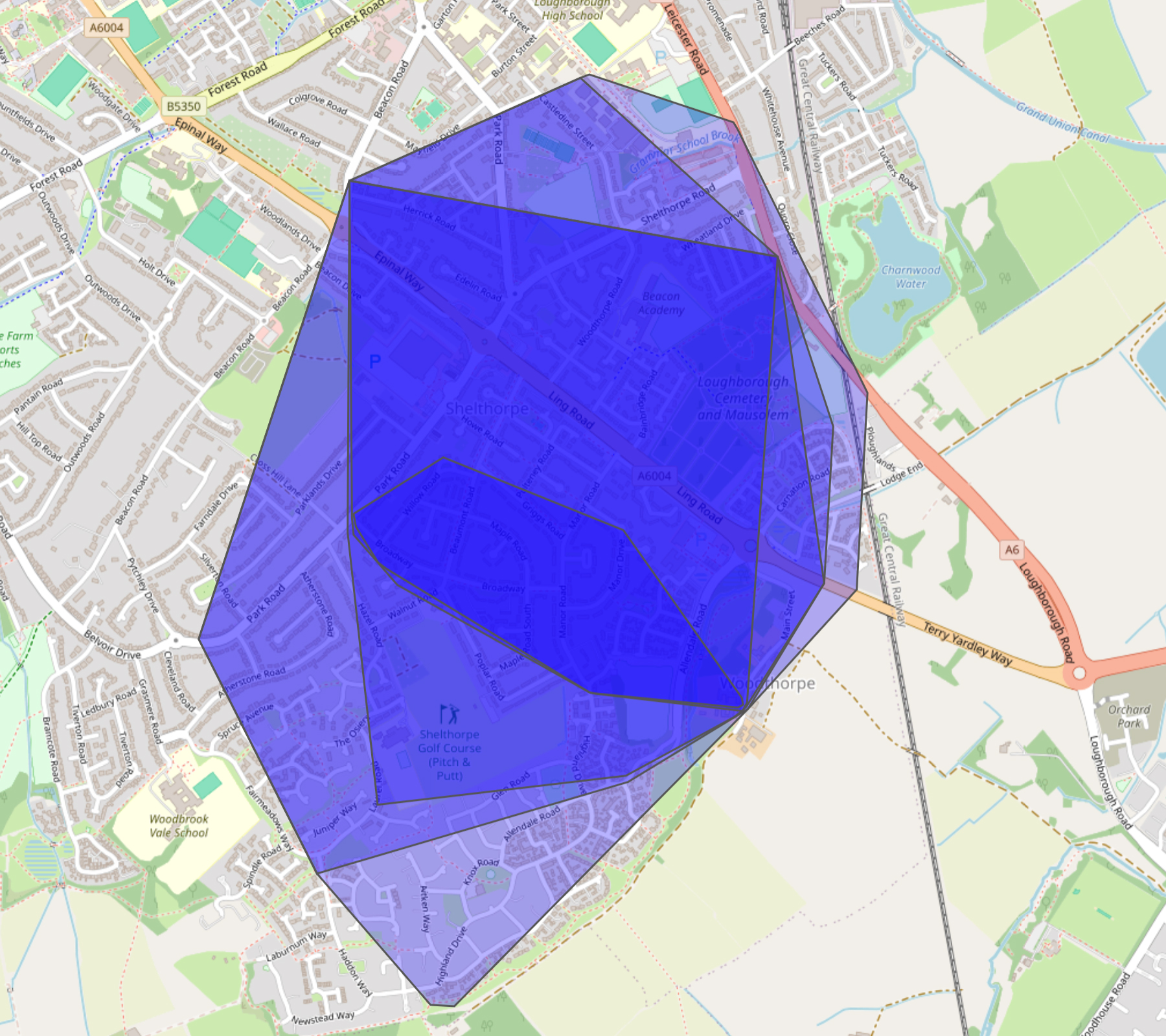}}
\subfloat[\centering]{\includegraphics[width=5.0cm]{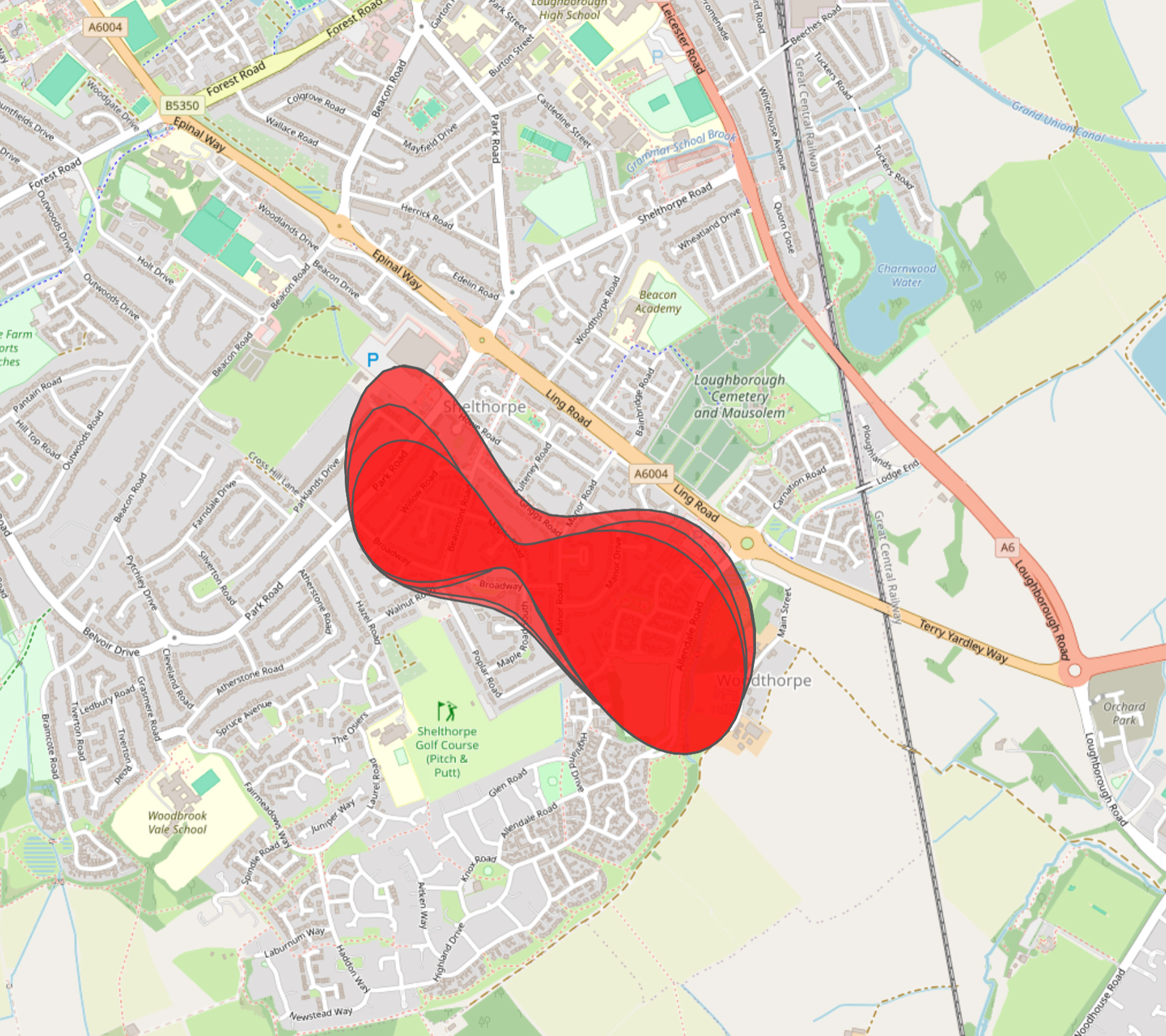}}\\
\end{adjustwidth}
\caption{Mobile signal coverage area of a single cell tower. On the left is the convex hull approach (\textbf{a}), which overgeneralises and includes large areas of no coverage. On the right is the OC-SVM approach, which captures concavities and internal holes to better represent the true coverage area. The right figure (\textbf{b}) is a smoothed, non-convex boundary that tightly wraps around the empirical coverage points, calculated using OC-SVM with RBF kernel.\label{fig2} \cite{OpenStreetMap}}
\end{figure} 


Meanwhile, the RBF width is governed by \( \gamma \) in \(K(x,x^{\prime}) = \exp(-\gamma\lVert x-x^{\prime} \rVert^2) \). A large \(\gamma\) yields a narrow influence radius, so each support vector only affects classification in its immediate vicinity and the learned decision boundary can become overly sensitive to noise. Conversely, a small \(\gamma\) produces a broad influence radius, allowing each support vector to influence a wide region and yielding a smoother, more general boundary that may underfit fine spatial detail. Selecting \(\gamma\) therefore sets the granularity of the learned surface.

To select the best hyperparameters, we tune \((\nu, \gamma)\) by temporal cross-validation using held-out time slices. We partition the data into training and validation sets along the temporal dimension (e.g., train on January measurements and validate on February). This simulates real-world deployment, where future coverage must be predicted from historical observations. For each candidate \((\nu, \gamma)\) pair, we train the OC-SVM on the training set and evaluate its performance on the validation set using metrics like recall (true positive rate) and precision (positive predictive value). We select the hyperparameters that yield the best trade-off between recall and precision on the validation set, ensuring robust generalisation to unseen data. 

In our approach, we partition the crowdsourced measurement data by signal levels and train a separate OC-SVM boundary for each level. The intuition is that the spatial support of good coverage points will differ (be more conservative) from that of weaker signal and so by modelling each level's support separately, we can obtain layered (nested) boundaries that demarcate zones of stronger versus weaker coverage. Each OC-SVM is responsible for estimating the region in which that signal level is reliably observed. In deployment, one can interpret a location's highest predicted signal level by querying which OC-SVM boundary it falls inside.

We compare our OC-SVM approach against the convex hull baseline by evaluating both methods on the same temporal validation splits. We assess their ability to correctly identify known coverage areas (true positives) and avoid falsely predicting coverage in known no-service areas (false positives). Metrics like recall, precision, and F1-score provide a quantitative comparison of their performance. We expect the OC-SVM to outperform the convex hull by better capturing complex, non-convex coverage boundaries and reducing false positives due to its learned, data-driven nature.

\section{Results}

To further understand the impact of hyperparameter tuning, we visualised the grid search results over a range of $(\nu, \gamma)$ values. The parameter $\nu \in (0, 1]$ serves as an upper bound on the fraction of training errors and a lower bound on the fraction of support vectors. Meanwhile, $\gamma$ controls the width of the radial basis function (RBF) kernel, thereby governing the smoothness of the decision boundary.

\begin{figure}[h]
\centering
\includegraphics[width=1.0\textwidth]{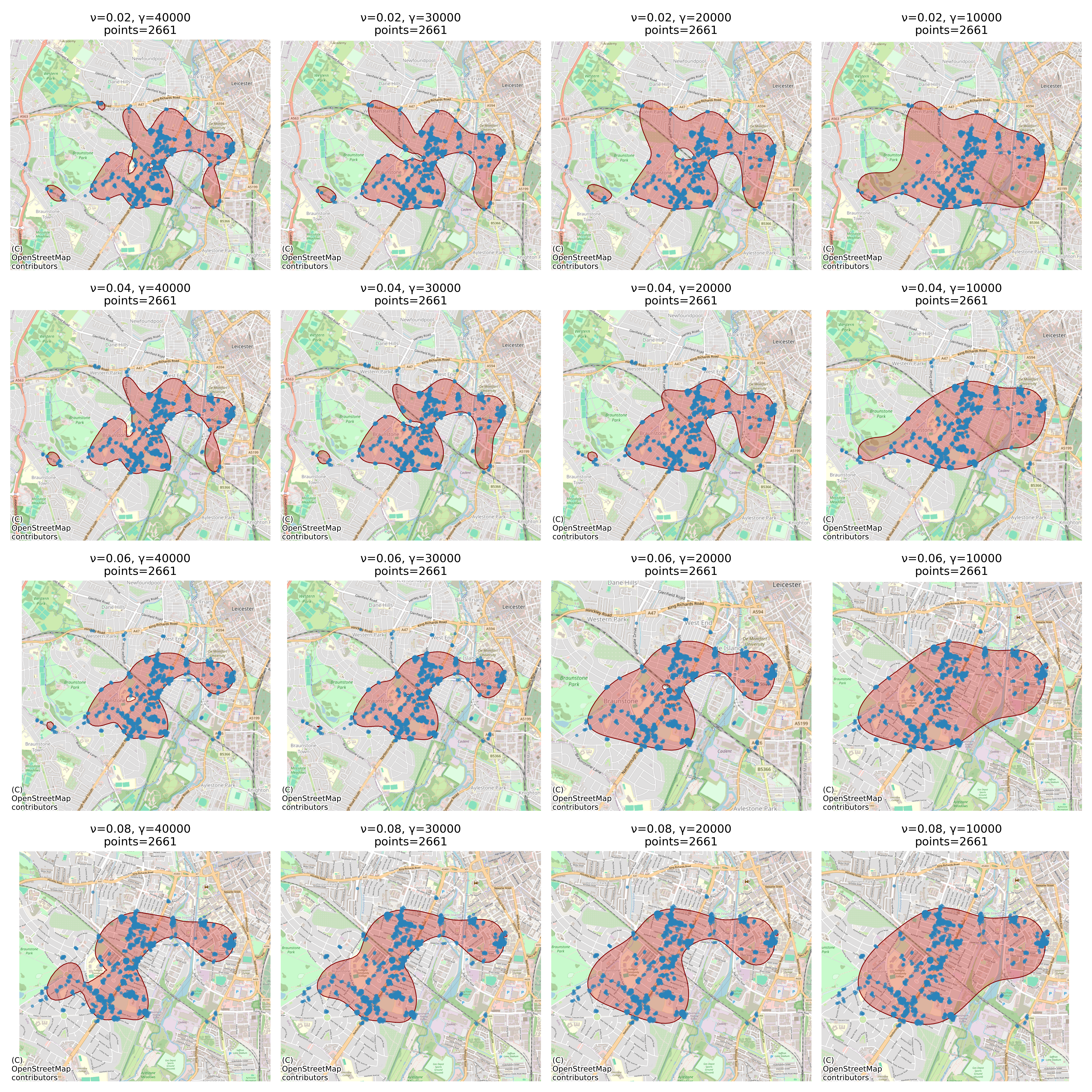}
\caption{Effect of varying hyperparameters $(\nu, \gamma)$ on the shape and fidelity of the OC-SVM boundary. Note the trade-off between smoothness and fragmentation.}
\label{fig:hyperparameter_grid}
\end{figure}

We conducted a grid search over $\nu \in \{0.02, 0.04, 0.06, 0.08\}$ and $\gamma \in \{1 \times 10^{4}, 2 \times 10^{4}, 3 \times 10^{4}, 4 \times 10^{4}\}$ using time-slice cross-validation. As illustrated in Figure \ref{fig:hyperparameter_grid}, lower values of $\nu$ produced tighter decision boundaries with fewer outliers, but often overfit the training data. In contrast, larger $\nu$ values allowed more points to lie outside the boundary.

The influence of $\gamma$ was equally significant. A small $\gamma$ value (e.g., $1 \times 10^{3}$) led to excessively smooth boundaries, which tended to underfit irregular coverage contours, particularly in urban microcells with complex signal topologies. Increasing $\gamma$ produced sharper boundaries, capable of capturing complex local fluctuations in signal strength. However, excessive $\gamma$ values (e.g., $\gamma=4 \times 10^{3}$) resulted in overfitting, where the decision surface became highly sensitive to local noise and yielded fragmented or disconnected coverage zones.

We systematically evaluate the performance of OC-SVM boundaries using various hyperparameters against the convex hull baseline. The choice of F1 score is particularly useful in our context because it provides a single metric that balances both precision and recall. This is important because, in coverage area estimation, we want to minimise both false positives (areas incorrectly marked as covered) and false negatives (missed coverage areas). Using the F1 score ensures that our model performs well across both metrics rather than optimising for one at the expense of the other. The F1 score formula is given by:

\[\text{F1} = 2 \cdot \frac{\text{Precision} \cdot \text{Recall}}{\text{Precision} + \text{Recall}}\]

\begin{table}[h]
\centering
\caption{Comparison of F1 Scores at different signal levels.}
\label{tab:results}
\small
\begin{tabular}{lccc}
\toprule
Category                           & Signal Level (dBm)      & Convex Hull  & OC-SVM \\
\midrule
1. Poor to none (outdoor only)     & \(<\) 105               & 0.140299     & 0.220113 \\
2. Variable (outdoor only)         & \(\geq\) -105 up to -95 & 0.090324     & 0.138441 \\
3. Good (outdoor only)             & \(\geq\) -95 up to -82  & 0.050785     & 0.073423 \\
4. Variable in-home, good outdoor  & \(\geq\) -82 up to -74  & 0.021470     & 0.027420 \\
5. Good in-home and outdoor        & \(\geq\) -74            & 0.011422     & 0.013407 \\
\bottomrule
\end{tabular}
\normalsize
\end{table}

\begin{figure}[h]
\centering
\includegraphics[width=0.6\textwidth]{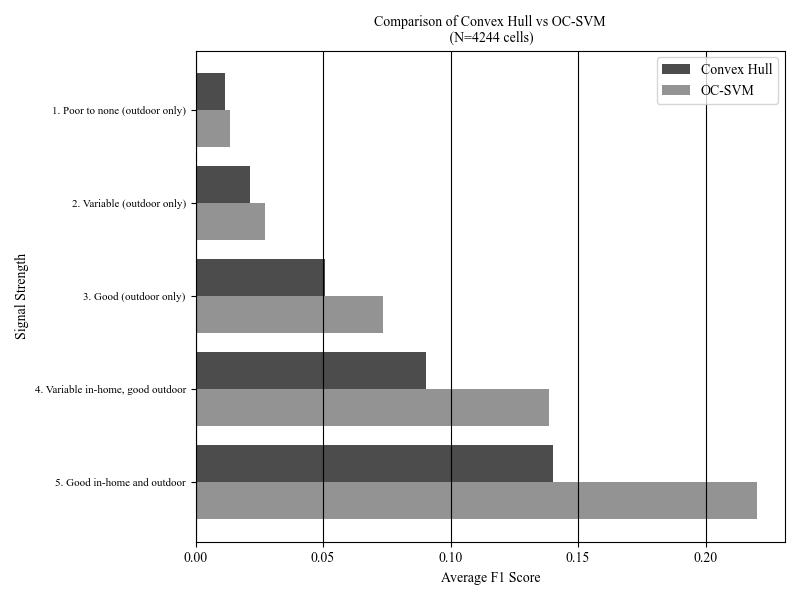}
\caption{F1 Scores of OC-SVM vs Convex Hull}
\label{fig:chart}
\end{figure}

We performed cross-validation by partitioning the data into training and validation sets along the temporal dimension. For each candidate pair of hyperparameters \((\nu, \gamma)\), we trained the OC-SVM on the training set and evaluated its performance on the validation set. The convex hull method was applied to the same training data for a fair baseline comparison. Each model was trained for individual cell towers to capture their unique coverage characteristics. We analysed more than 4,000 cell towers across England for our investigation, ensuring a diverse representation of urban and rural environments. The results in Table \ref{tab:results} and Figure \ref{fig:chart} are averaged over all cell towers. We observed that the OC-SVM approach consistently outperforms the convex hull approach, particularly in scenarios with variable or poor signal levels where coverage boundaries are more complex. This demonstrates the effectiveness of the OC-SVM in capturing non-convex coverage areas and reducing false positives.

\section{Discussion}

This study set out to address the challenge of accurately estimating mobile network coverage areas using crowdsourced data. The primary aim was to evaluate whether the OC-SVM could provide more precise and reliable coverage boundaries compared to a more traditional geometric method using a convex hull. By doing so, we sought to improve the identification of coverage gaps and weak spots, ultimately supporting better network planning and QoE.

The results demonstrate that the OC-SVM approach consistently improves on the accuracy of coverage boundary estimation compared to the convex hull approach. This is particularly significant in urban environments where coverage gaps are often hidden by complex terrain and building structures. The kernelised decision boundary captures complex, non-convex shapes that more accurately reflect real-world coverage, improving precision because the model is less likely to include areas without service and enhancing recall because it better identifies true coverage areas.


Numerous attempts have utilised Machine Learning (ML) techniques to enhance traditional path loss models \cite{path_loss_ml, 5g_rf_model, mlp_path_loss}. However, applying an OC-SVM classifier to crowdsourced data for coverage estimation is a novel contribution that extends the use of ML in wireless communications. This approach draws on ML's strength in handling complex, non-linear relationships, making it well suited to capturing the intricacies of mobile network coverage that traditional methods may overlook. Incorporating crowdsourced data also provides an empirical basis for understanding real-world coverage patterns, rather than relying solely on theoretical models or limited drive-test data that are both expensive and time-consuming to collect. Additionally, the utilisation of crowdsourced measurements allows for a dynamic and up-to-date representation of network coverage, reflecting evolving changes in the environment and user behaviour that static models may miss, such as new building developments, changes in user density, and seasonal variations in the environment.

We assume that the crowdsourced measurements are sufficiently representative of the underlying spatial distribution of coverage, despite the likelihood of uneven sampling density across different regions. In practice, areas with higher population density or greater user activity may be overrepresented relative to rural or low-traffic areas.

The findings suggest that the OC-SVM model is better equipped to handle the complexities of real-world mobile network coverage. By leveraging the strengths of machine learning, particularly in identifying patterns and anomalies in data, the OC-SVM can provide a more nuanced understanding of coverage areas. This has the potential to inform more effective network design and optimisation strategies.

The improved accuracy of coverage boundary estimation has significant implications for network planning and optimisation. By accurately identifying coverage gaps and weak spots, operators can make informed decisions about where to deploy additional resources to address service gaps, such as new cell sites or signal boosters, and to select appropriate spectrum bandings for both rural and urban areas. This targeted approach can lead to more efficient use of resources and ultimately enhance the QoE for end users.

\section{Conclusion and Future Work}

This study proposes the OC-SVM as a method for mobile network coverage analysis using crowdsourced geolocation data. We compared the proposed approach against a geometric convex hull baseline and demonstrated that kernelised OC-SVM can effectively capture complex, non-convex coverage boundaries that better reflect real-world signal propagation, particularly in urban environments with significant clutter, terrain, and shadowing effects. With appropriate tuning, the SVM boundaries achieve a balance between coverage inclusivity and the exclusion of spurious outliers, enabling the estimation of coverage boundaries at various signal strength levels. The results confirm that crowdsourced data, despite inherent heterogeneity and noise, can be transformed into reliable coverage models when combined with suitable machine learning techniques. This offers direct value both to operators seeking to prioritise investment and to regulators monitoring equitable service provision.

Looking ahead, several avenues remain open for extension and refinement. First, comparative studies across multiple network operators would enable systematic benchmarking of coverage quality, competition, and consumer choice, thereby extending the present analysis to the policy domain. Second, integrating external geographic layers such as digital elevation models, building footprints, or clutter maps could improve interpretability by directly linking weak coverage spots to underlying physical obstructions.

Methodological extensions also offer promising opportunities. Hybrid models that combine One-Class SVM with other machine learning or geostatistical approaches could provide richer representations. For example, Gaussian Processes \cite{Rasmussen2005} or Random Forest regressors could be employed to capture probabilistic uncertainty, while classical Kriging techniques \cite{Cressie1993} could improve the imputation of missing data. Likewise, geometric refinements such as $\alpha$-shapes \cite{Edelsbrunner1983} provide an alternative to convex hulls, allowing coverage boundaries that better adapt to concave and irregular geometries. 


Because kernelised OC-SVM natively supports multi-dimensional inputs, the proposed approach can be extended to 3D geolocation data (i.e., longitude, latitude, altitude). This could further enrich coverage analysis, particularly in dense urban environments where verticality plays a key role.

Finally, extending this framework to capture temporal and technological dimensions would yield significant value. Continuous monitoring could reveal dynamic coverage fluctuations, such as congestion, seasonal effects, or post-deployment improvements. Applying the methodology across multiple spectrum bands, technologies (e.g. 4G, 5G) or handsets would enable comparative analyses of network performance and technology evolution.

In summary, the proposed method demonstrates that a machine-learning approach can be used to estimate mobile coverage boundaries from empirical crowdsourced data. This opens up new possibilities for network analysis, optimisation, and regulation. With further refinement and extension, this framework has the potential to become a standard tool for mobile network coverage assessment in the era of machine learning.

\appendixtitles{no} 
\appendixstart
\appendix
\section[\appendixname~\thesection]{}
The dataset used in this study originates from NetPerform \cite{vodafone_netperform}, Vodafone's crowdsourced measurement system for assessing empirical mobile network performance. NetPerform collects network experience data directly from mobile handsets via an embedded software development kit (SDK) integrated within user applications. The SDK passively records connectivity parameters during normal device usage, providing a large-scale, user-centred view of network performance.

Each measurement record includes information such as the timestamp, geographic coordinates, radio technologies (2G, 3G, 4G, and 5G), and key performance indicators including received signal strength, throughput, latency, and service availability. These metrics reflect the network's performance as experienced by users, rather than being inferred from network-side counters. All data are anonymised, aggregated, and processed in compliance with data protection and privacy frameworks before being made available for analysis.

For this study, a synthetically generated dataset based on UK NetPerform was used \cite{netperform_synthetic}. The synthetic data comprise geolocated mobile signal measurements spanning multiple months, capturing a broad range of environmental and temporal conditions. The synthetic dataset underwent standard pre-processing, including coordinate validation, outlier removal, and spatial filtering to exclude duplicate events.

Unlike traditional drive-test datasets, which are constrained to predefined routes and time periods, NetPerform provides continuous and naturally distributed coverage driven by customer behaviour. This allows for a more representative and granular analysis of empirical network experience across urban, suburban, and rural environments.

\dataavailability{
	The synthetic dataset used in this study is available at \url{https://huggingface.co/datasets/joefee/cell-service-data}. The dataset comprises geolocated mobile signal measurements spanning across multiple months, capturing a broad range of environmental and temporal conditions.
}

\reftitle{References}


\bibliography{bibliography.bib}

\begin{thebibliography}{999}

\bibitem[Ofcom(2024{\natexlab{a}})]{ofcom_drive_test}
Ofcom.
\newblock Mobile signal strength measurement systems,  2024.

\bibitem[Ofcom(2024{\natexlab{b}})]{ofcom_drive_route}
Ofcom.
\newblock Drive route map,  2024.

\bibitem[Koutroumpis and Leiponen(2016)]{Koutroumpis2016}
Koutroumpis, P.; Leiponen, A.
\newblock Crowdsourcing mobile coverage.
\newblock {\em Telecommunications Policy} {\bf 2016}, {\em 40},~532–544.
\newblock {\url{https://doi.org/10.1016/j.telpol.2016.02.005}}.

\bibitem[Neidhardt et~al.(2013)Neidhardt, Uzun, Bareth, and Kupper]{Neidhardt2013}
Neidhardt, E.; Uzun, A.; Bareth, U.; Kupper, A.
\newblock Estimating locations and coverage areas of mobile network cells based on crowdsourced data.
\newblock In Proceedings of the 6th Joint IFIP Wireless and Mobile Networking Conference (WMNC). IEEE,  Apr 2013, p. 1–8.
\newblock {\url{https://doi.org/10.1109/wmnc.2013.6549010}}.

\bibitem[Brunello et~al.(2023)Brunello, Dalla~Torre, Gallo, Gubiani, Montanari, and Saccomanno]{Brunello2013}
Brunello, A.; Dalla~Torre, A.; Gallo, P.; Gubiani, D.; Montanari, A.; Saccomanno, N.
\newblock Crowdsourced Reconstruction of Cellular Networks to Serve Outdoor Positioning: Modeling, Validation and Analysis.
\newblock {\em Sensors} {\bf 2023}, {\em 23}.
\newblock {\url{https://doi.org/10.3390/s23010352}}.

\bibitem[Schölkopf et~al.(2001)Schölkopf, Platt, Shawe-Taylor, Smola, and Williamson]{10.1162/089976601750264965}
Schölkopf, B.; Platt, J.C.; Shawe-Taylor, J.; Smola, A.J.; Williamson, R.C.
\newblock Estimating the Support of a High-Dimensional Distribution.
\newblock {\em Neural Computation} {\bf 2001}, {\em 13},~1443--1471,  \href{http://arxiv.org/abs/https://direct.mit.edu/neco/article-pdf/13/7/1443/814849/089976601750264965.pdf}{{\normalfont [https://direct.mit.edu/neco/article-pdf/13/7/1443/814849/089976601750264965.pdf]}}.
\newblock {\url{https://doi.org/10.1162/089976601750264965}}.

\bibitem[Ope(2025)]{OpenStreetMap}
Planet dump retrieved from https://planet.osm.org,  2025.

\bibitem[Zhang et~al.(2019)Zhang, Wen, Yang, He, and Wang]{path_loss_ml}
Zhang, Y.; Wen, J.; Yang, G.; He, Z.; Wang, J.
\newblock Path Loss Prediction Based on Machine Learning: Principle, Method, and Data Expansion.
\newblock {\em Applied Sciences} {\bf 2019}, {\em 9},~1908.
\newblock {\url{https://doi.org/10.3390/app9091908}}.

\bibitem[Sousa et~al.(2021)Sousa, Alves, Vieira, Queluz, and Rodrigues]{5g_rf_model}
Sousa, M.; Alves, A.; Vieira, P.; Queluz, M.; Rodrigues, A.
\newblock Analysis and Optimization of 5G Coverage Predictions Using a Beamforming Antenna Model and Real Drive Test Measurements.
\newblock {\em IEEE Access} {\bf 2021}, {\em 9},~101787--101808.
\newblock {\url{https://doi.org/10.1109/ACCESS.2021.3097633}}.

\bibitem[Isabona et~al.(2022)Isabona, Imoize, Ojo, Karunwi, Kim, Lee, and Li]{mlp_path_loss}
Isabona, J.; Imoize, A.; Ojo, S.; Karunwi, O.; Kim, Y.; Lee, C.C.; Li, C.T.
\newblock Development of a Multilayer Perceptron Neural Network for Optimal Predictive Modeling in Urban Microcellular Radio Environments.
\newblock {\em Applied Sciences} {\bf 2022}, {\em 12}.
\newblock {\url{https://doi.org/10.3390/app12115713}}.

\bibitem[Rasmussen and Williams(2005)]{Rasmussen2005}
Rasmussen, C.E.; Williams, C.K.I.
\newblock {\em Gaussian Processes for Machine Learning}; The MIT Press,  2005.
\newblock {\url{https://doi.org/10.7551/mitpress/3206.001.0001}}.

\bibitem[Cressie(1993)]{Cressie1993}
Cressie, N.A.C.
\newblock {\em Statistics for Spatial Data}; Wiley,  1993.
\newblock {\url{https://doi.org/10.1002/9781119115151}}.

\bibitem[Edelsbrunner et~al.(1983)Edelsbrunner, Kirkpatrick, and Seidel]{Edelsbrunner1983}
Edelsbrunner, H.; Kirkpatrick, D.; Seidel, R.
\newblock On the shape of a set of points in the plane.
\newblock {\em IEEE Transactions on Information Theory} {\bf 1983}, {\em 29},~551--559.
\newblock {\url{https://doi.org/10.1109/TIT.1983.1056714}}.

\bibitem[Vodafone(2025)]{vodafone_netperform}
Vodafone.
\newblock Vodafone NetPerform,  2025.

\bibitem[Feehily et~al.(2025)Feehily, Freeman, and Wong]{netperform_synthetic}
Feehily, J.; Freeman, T.; Wong, T.
\newblock Synthetic Mobile Network Performance Dataset,  2025.
\newblock {\url{https://doi.org/10.57967/hf/6654}}.

\end{thebibliography}

\PublishersNote{}
\end{document}